# Networked Intelligence

Towards Autonomous Cyber Physical Systems

André Karpištšenko
andre@networked.ai

Abstract

Developing intelligent systems requires the combination of results from both industry and academia. In this report you find an overview of relevant research fields and industrially applicable technologies for building large scale cyber physical systems. A conceptual architecture is used to illustrate how existing pieces may fit together, and the maturity of the subsystems is estimated.

The goal is to structure developments and the challenge of machine intelligence for Consumer and Industrial Internet technologists, cyber physical systems researchers and people interested in the convergence of data and the Internet of Things (IoT). This report will be useful for planning developments of intelligent systems.

*Acknowledgements:* I would like to thank Ando Saabas for help in finding the focus, Lauri Koobas and Erki Suurjaak for pushing for brevity and clarity, Meelis Kull, Tambet Matiisen, Anna Leontjeva, Taivo Pungas and Kairit Sirts for improvements. I'd also like to thank Machine Learning Estonia presenters for providing the information on the latest state in machine learning, and Skype ML and AI reading group for updates in the field. I am grateful to the early readers as well.

―――――

Technical Report, 2016

# Executive Summary

The benefits of intelligent systems are demonstrated by many artificial narrow intelligence (ANI) applications used by businesses, governments and society. Some are highlighted later in the Applications section. In developing intelligent systems there is a positive feedback loop - as the products, services and processes get better, the increase in profitability is reinvested into further developments. There are many more innovations one can imagine. Only what exists on the market today is highlighted here.

Cyber physical systems integrate computational and physical capabilities to interact with humans, machines, and the environment. This report brings together the progress in developments of such systems to develop a vision of a concept architecture for the machine intelligence challenge. To demonstrate the feasibility of the concept we give an overview of relevant existing research results and technologies necessary for building it.

The report is structured as follows. In the Research Progress the focus is mainly on progress made in computer science, while highlighting some of the adjacent fields. The Technology Action provides an overview of trends in data and IoT. Core infrastructure such as data & IoT platforms and advanced analytics platforms are maturing with potential high-value commercial applications. Relevant to the intelligent systems, the highest maturity is in data gathering and preprocessing, followed by advanced analytics.

In the illustrative section How the Pieces Fit Together, a hierarchical system of systems approach is used, consisting of parts related to data preprocessing, models and self-improvement. Higher-order functions related to finding solutions with limited resources[1] have been left out, while the aspects where the involvement of people is necessary are highlighted. This vision is not intended as a technical blueprint. I estimated that the concept illustrated here is achievable within a 5-10 year time frame.

Bringing together research and technology developments is necessary for engineering intelligent systems. Hopefully this report will spark developments and conversations that speed up the progress.

The report is intended for Consumer and Industrial Internet technologists, cyber physical systems researchers and people interested in the convergence of data and the IoT. It will be invaluable for planning the development of intelligent systems, identifying bottlenecks, estimating resources and selecting technologies.

***How to read guide****:* The first sections give the context for the vision. If you are practically minded, start from Technology Action. If you know the market and seek ideas for your system, start from the section How the Pieces Fit Together.

---

[1] R. Kurzweil, How to Create a Mind: The Secret of Human Thought Revealed, Penguin Books, 2012

# Research Progress

*The aim of this section is to describe the research fields directly relevant to developing and designing large scale autonomous cyber physical systems[2].*

One of the foundations for developing large scale systems is the system of systems concept. That is, large scale integrated systems which are diverse and independently operable on their own, but are networked together for a common goal.[3]

The field of artificial general intelligence (AGI)[4] is the forefront of related research fields. The approach here is pragmatic and business driven. Given the possible long road to practical AGI, there is no focus here on providing a complete overview. In the industry, combinations of ANIs are used to solve complex situations.

The Figure 1 below illustrates how different research fields fit together to form networked systems. It is assumed that creating intelligent systems is mainly a software and systems engineering challenge, thus fields like mechanical and electrical engineering, which focus on physical objects and their applications, are excluded. However, some of the relevant results from these fields are listed in the Appendix.

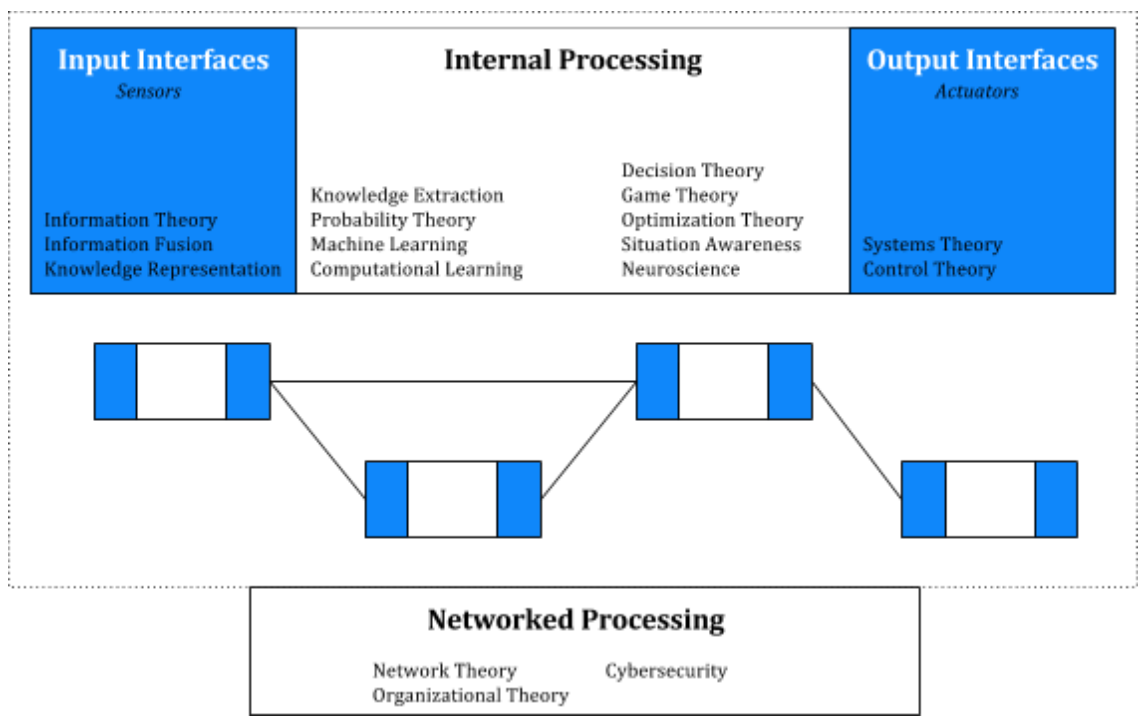

*Figure 1. Illustration of relevant research fields needed the for design and development of networks of systems*

In Figure 1, research is grouped into one of four categories depending on the key aspect they address. In the following sections you will find a brief overview and highlights with links to technical material.

---

[2] Partially inspired by IARPA Research Programs
[3] M. Jamshidi, Systems of Systems Engineering: Principles and Applications, CRC Press, 2008
[4] P. Wang, B. Goertzel, Theoretical Foundations of Artificial General Intelligence, Atlantis Press, 2012

## Input Interfaces

The fields most relevant to the systems' ability to take input from external systems and to observe context and environment are information theory, information fusion[5] and knowledge representation[6].

Information theory defines the limits of systems coping with exponentially increasing[7] amount of incoming data in processing and storage infrastructure. Information fusion deals with the art of merging information from heterogeneous sources. For cyber physical systems, a relevant subset of the field is sensor fusion[8]. It is becoming commonplace that information fusion is a stream processing activity that operates with continuous data flows. In modern data processing and analysis systems, multimodal information integration is the most time consuming activity. This is due to high variety of data formats, versioning of data and variety in content. Successful fusion relies on data curation and domain knowledge that can be partially automated. Knowledge representation is about information required to understand incoming data. The extra structural, semantic and meta information allows machines to process raw data into something meaningful.

There are different types of data that cyber physical systems transform into a knowledge-base: textual data, usually in natural language, but also in formal languages; image data, for example photographs, satellite imagery and other; video data; audio data; numeric data and other binary signals. At large scales, a key aspect is to provide means to cope with large datasets which may also contain errors.

## Internal Processing

The following research fields aim at making sense of the incoming and historic information and internal system state to find patterns, make predictions and take decisions to deliver solutions for applications.

The basis is to make stored data, information and knowledge available in machine-readable and machine-interpretable formats. Indexing and probabilistic retrieval methods decide what data can be efficiently accessed and retrieved within query results. Information extraction and retrieval[9] is a field that has evolved rapidly after the onset of web era and is highly relevant to working with text, audio transcripts and image & video captions. Knowledge organization and validation related functionalities are further layers on top of related technologies.

Probability theory that form the foundations for many aspects in intelligent systems, such as information retrieval, machine learning (ML) and knowledge discovery, requires highlighting. Probabilistic programming languages[10] focus on simplifying software engineering that deals with probabilities, so that less code needs to be written. Bayesian inference and subjective Bayesian probability form an important basis for designing models capable of dynamic problem solving.

---

[5] V. Torra, Information Fusion in Data Mining, Springer, 2003
[6] R. Brachman, H. Levesque, Knowledge Representation and Reasoning, Morgan Kaufmann 2004
[7] EMC Digital Universe with Research & Analysis by IDC
[8] Mitchell, H. B. Multi-sensor Data Fusion, Springer-Verlag, 2007
[9] C. Manning, P. Raghavan, H. Schütze, Introduction to Information Retrieval, Cambridge University Press, 2008
[10] http://probabilistic-programming.org/

## Machine Learning and Knowledge Discovery

Machine Learning[11] is a rapidly advancing field with a range of different statistical learning methods[12] which can be used for supervised, partially supervised and unsupervised learning. A method quickly gaining traction is deep learning[13] which is about representation learning and automated feature engineering. The branch is used in many industrial applications. Other popular methods are random forests and gradient boosting.

An increasing trend is one of hybrid intelligent systems where multiple methods are combined to benefit from the strengths of each[14]. The approach combines neural networks, expert systems, fuzzy logic, symbolic systems, genetic algorithms and case-based reasoning.

A common aspect of machine learning is its dependence on data. There exists a subset of intelligently solvable problems that require less data, like those based on chemical and physical processes. Such models are based on systems of differential equations and laws of nature. There is a trend of replacing some classical algorithms with neural network based approximations to improve performance[15].

In developing intelligent systems one must choose between models as well as create ensemble models. To ease the process of communication, models should be interpretable. See the example of random forests[16].

## Computational Learning and Process Automation

Computational learning theory deals with the design and analysis of machine learning algorithms, mainly concerned with proving algorithms. Here a broader approach is taken. Self-improvement processes[17] and the applications of machine learning to guide model development are not new. For example, neuroevolution uses evolutionary algorithms for training[18] and online machine learning deals with continuous sequential data, updating the model at each step. As more models are produced, model curation and deployment automation is necessary.

Development of new models involves data preparation steps such as normalization, semi-automated feature extraction and dimensionality reduction. To perform data exploration for useful pattern discovery in high-dimensional data, methods such as topological data analysis[19] can be used. Some progress has been made in automating feature relationship identification[20].

## Other Fields

There are other important fields, such as data mining, that are not covered here. In building cyber physical systems there are more aspects than just working with the data and models. For example to decide on actions, statistical decision theory facilitates identification of decision input uncertainties and finding a solution. To deal with optimally unsolvable situations,[21] heuristics can be used. Attention based methods invest resources based on areas of interest[22].

---

[11] K. Murphy, Machine Learning: A Probabilistic Perspective, MIT Press, 2012
[12] T. Hastie, R. Tibshirani, J. Friedman, The Elements of Statistical Learning: Data Mining, Inference, and Prediction, Second Edition, Springer, 2009
[13] I. Goodfellow, Y. Bengio and A. Courville, Deep Learning, MIT Press (in preparation), 2016
[14] L. Medsker, Hybrid Intelligent Systems, Springer, 2013
[15] K. Abhishek, M.P. Singh, S. Ghosh, A. Anand, Weather Forecasting Model using Artificial Neural Network, C3IT, 2012
[16] A. Saabas, Interpreting random forests, Diving into data blog, 2014
[17] J. Dunn, Introducing FBLearner Flow: Facebook's AI backbone, Facebook Code, 2016
[18] D. Floreano, P. Dürr, C. Mattiussi, Neuroevolution, Evolutionary Intelligence, 2008
[19] G. Carlsson, Topology and Data, Bulletin of the American Mathematical Society, 2009
[20] D. Reshef, Y. Reshef, H. Finucane, S. Grossman, G. McVean, P. Turnbaugh, E. Lander, M. Mitzenmacher, P. Sabeti, Detecting novel associations in large datasets, Science, 2011

In industries such as finance, where systems are in co-opetition, means to think strategically are built using game theory[23]. While this aspect is not relevant for finding a solution in a one off application, it becomes relevant when operating continuously. Finding solutions for useful situations requires explicit accounting for internal and external resource constraints. Optimization theory deals with selecting the best element from a set of available alternatives. In machine learning, regularization deals with overfitting.

For autonomous systems, situation awareness that originated in human factors research has recently advanced to hierarchical situation awareness for cyber physical systems[24]. This is relevant to perceiving context and comprehending it.

Neuroscience[25], especially computational neuroscience, provides inspiration for ways of structuring intelligent systems. Progress in neuroscience[26] can be applicable in creating autonomous cyber physical systems. Initial progress has been made with robots[27].

## Output Interfaces

For interacting with external systems and the environment, general system theory has broad applicability. The most relevant class is cyber physical systems[28] that refers to systems with integrated computational and physical capabilities that interact with humans and machines.

Systems have to deal with dynamic behaviour and control theory provides guidance for process control in order to steer a system towards a desired state. Many AI control challenge concepts and behaviours are manifested in simple setups[29]. Control interfaces can take many forms, especially for human-machine interaction, for example natural language interfaces. Human-machine interaction is an essential part of an intelligent system. Extended intelligence[30] is an emerging field studying intelligence as a fundamentally distributed phenomenon.

## Networked Processing

System of systems forms a dynamically changing interactive network of processes that encode or decode information from one form to another. Network theory provides a means to understand the form and function of the system that the network represents[31]. Utilizing network theory, system designers can start to reason about its structure, network resilience and resistance to perturbations. Advancements in control of networks [32] provide an initial framework for controlling complex self-organized systems. More specifically, one can identify the set of driver nodes with time-dependent control to guide the system's entire dynamics. However, sparse inhomogeneous networks, which emerge in many real complex systems, are the most difficult to control. In addition networked systems may be represented as multilayer structures[33], even evolving as a function of time[34]. Methodologies dealing with such representations sets the contemporary directions in network science with straightforward applications in system design, IoT, and interdependent systems. Knowledge about the network can be classified and visualized for designers and operators[35].

---

[21] C. Moore, S. Mertens, The Nature of Computation, Oxford University Press, 2011
[22] A. Almahairi, N. Ballas, T. Cooijmans, Y. Zheng, H. Larochelle, A. C., Dynamic Capacity Networks, ICLR 2016
[23] S. Hargreaves Heap, Y. Varoufakis, Game Theory: A Critical Introduction, Routledge, 1995
[24] J. Preden, Enhancing Situation-Awareness, Cognition and Reasoning of Ad-Hoc Network Agents, Tallinn University of Technology, 2010
[25] Human Brain Project, EU FET Flagship Initiative
[26] B. Baars, Global workspace theory of consciousness: toward a cognitive neuroscience of human experience, Progress in Brain Research, 2005
[27] S. Bringsjord, J. Licato, N. Govindarajulu, R. Ghosh, A. Sen, Real Robots that Pass Human Tests of Self-Consciousness, IEEE Ro-Man, 2015
[28] R. Baheti, H. Gill, Cyber-physical Systems, The Impact of Control Theory, IEEE Control Systems Society, 2011
[29] J. Tallinn, Toy Model of the Control Problem, 2015, http://www.slideshare.net/AndrKarpitenko/ai-control
[30] http://www.pubpub.org/pub/extended-intelligence
[31] M. Newman, Networks: An Introduction, Oxford University Press, 2010
[32] Y. Liu, J. Slotine, A. Barabasi, Controllability of Complex Networks, Nature, 2011

Drawing the parallel between a dedicated system and an agent, a traditional multi-agent systems approach provides concepts for structuring organizations of agents that interact with the environment. Using organizational theory as inspiration for organizing a cyber physical system of systems (for example the fields of self-organization, scientific management and organizational processes) can aid designers in choosing the right protocols for evolving their systems.

Risks emerge when a system interacts with people and other systems. Cybersecurity is about managing these risks with appropriate measures. For example cryptographic methods can be used for exchanging sensitive information. Information security is a cross-cutting concern and an important aspect of any system, especially so for intelligent systems interacting with people and nature. Intelligent system design should factor in threat intelligence and modeling to account for the risks of its actions and operations.

## Technology Action

*An overview of state of the art in technologies relevant to large scale autonomous cyber physical systems[36].*

These are fun times for anyone in software and systems engineering. The systems are becoming intelligent, boundaries between fields are blurring and the pace of development is increasing. Mature trends like mobile and new exponential trends like Industrial IoT, Consumer IoT, natural interfaces and data science are creating new commercial and technological opportunities. The list of technologies and companies is long, here a summary of some promising developments is provided.

---

[33] Kivelä, M., Arenas, A., Barthelemy, M., Gleeson, J.P., Moreno, Y. and Porter, M.A., 2014. Multilayer networks. Journal of complex networks, 2(3), pp.203-271.
[34] Holme, P. and Saramäki, J., 2012. Temporal networks. Physics reports, 519(3), pp.97-125.
[35] K. Börner, Atlas of Science: Visualizing What We Know, MIT Press, 2010
[36] Inspired by market research done by Shivon Zilis and FirstMark Capital

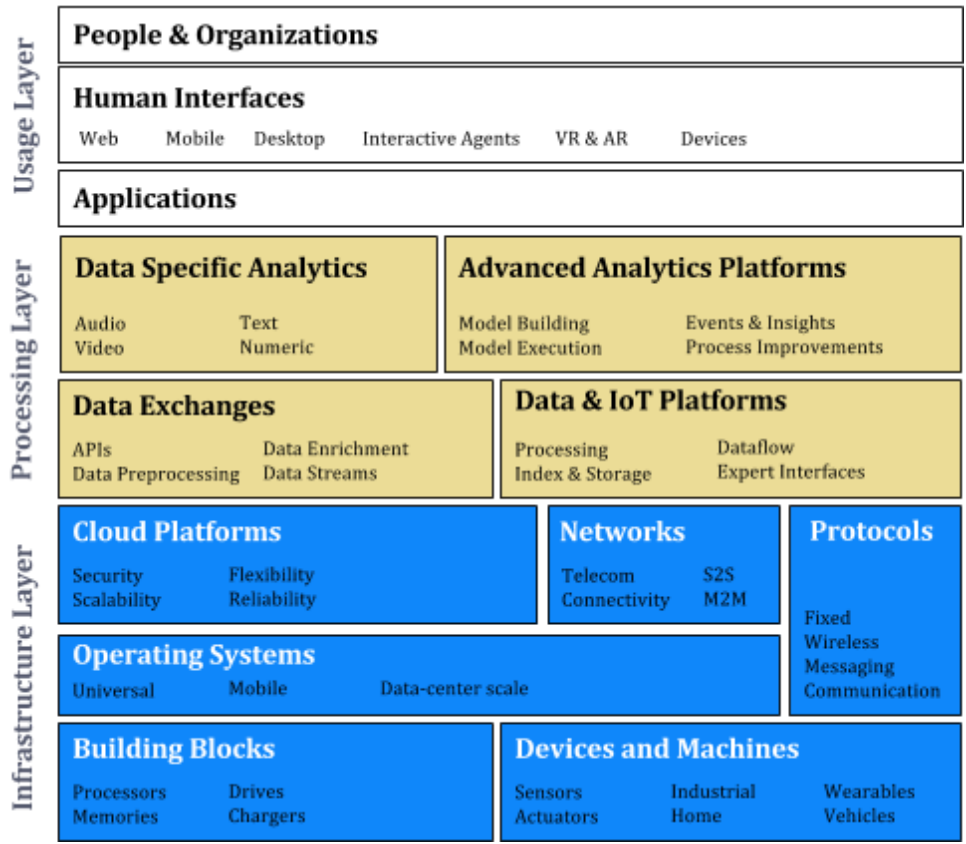

*Figure 2. Technology areas relevant to designing and building intelligent systems*

The technologies relevant to building cyber physical systems are grouped into three categories: Infrastructure Layer, Processing Layer and Usage Layer as illustrated in Figure 2. In the following sections you will find an overview of the categories[37].

## Infrastructure Layer

The infrastructure offers basic building blocks for creating commercially viable cyber physical systems. The technologies define what is achievable in meeting non-functional requirements.

Details on the overview of the Infrastructure Layer are explained in the Appendix. The base infrastructure exists for development of autonomous cyber physical systems and is advancing fast. The Data & IoT Platforms, as well as Data Exchanges address the need to adapt to the increasing amount of devices, machines and protocols.

## Processing Layer

The processing layer delivers the basis for building the core functionality of intelligent systems.

---

[37] For paper readers, link to companies and technologies are provided, use the links to access the highlights directly.

Data & IoT Platforms

Data and devices are the cornerstones of cyber physical systems. The available datasets and data streams define the scope of applications. Processing, storing, indexing and querying the data (e.g. Drill) are the most fundamental functions.

Data platforms address working with data, companies like Cloudera (recently launched Kudu), MapR and others provide open source and proprietary technologies that scale to very large data sizes. Company products like HortonWorks HDF further extend these technologies with IoT ready solutions that integrate open-source technologies from organizations like LinkedIn (Kafka), Twitter (Storm) and NSA (Nifi). Further progress is made by companies like DataBricks (Spark), Data Artisans (Flink) and others who commercialize open-source research results of thought leaders in data systems like UC Berkeley (AMPLab), MIT (DBg), CMU (DB Group) and other institutions. A driving trend is the move from general purpose data warehouses and relational databases to purpose built databases for small-data, columnar, time-series, array, key-value and other types data. Relevant to intelligent systems, probabilistic platforms are emerging. Cloud platform providers commoditize the most widely used data solutions for petabyte scale needs.

While the solutions for big data are maturing and converging, the early phase of the IoT trend has seen a response by many different companies like Cisco Jasper and PTC ThingWorx who are competing for the leading position in different markets. Driven by the data and IoT trends, traditional companies like GE are redefining themselves as software-driven service companies and are strategically investing into partnerships [38]. Companies such as Pivotal are recognizing this major shift and deliver software for those enterprises. Younger companies for example Helium and others, are addressing IoT development needs to speed up the time it takes for new connected devices to reach mass markets. Many IoT development platforms, like Hologram, focus on the unique needs that come with dealing with many sensors, devices and data streams.

Data and IoT trends converge in platforms like Planet OS that combine open-source, proprietary and cloud technologies to deliver real value in demanding industries like energy. Reactive platforms like Lightbend form a good basis for building such platforms. Building upon such endeavours brings the benefit of machine and sensor independent data formats and protocols.

Data Preprocessing and Exchanges

Finding the right data, cleaning, curating and enriching it is one of the most time-consuming activities in data science. An intelligent system depends on the quality of input directly. Thus it makes sense to rely on dedicated technologies like Tamr, Trifacta and CrowdFlower in these activities. For data integration, platforms like snapLogic and others exist.

Preprocessing endeavours are commonly too large for a single entity. A solution is delivered by data syndication and publishing services, such as Xignite, Planet OS, Enigma, Qlik and potentially Microsoft Azure DataMarket, that provide high quality datasets and data streams via well-defined APIs. Building upon Data Exchanges and preprocessing related technologies is a good use of resources.

---

[38] https://www.oracle.com/corporate/pressrelease/ge-digital-042116.html

## Advanced Analytics Platforms

Advanced analytics is the part of Technology Action dealing directly with putting intelligence into systems. Given high quality input, model building, execution, optimization and postprocessing are the major steps in analytics. The major drivers are the need for increasing the speed of learning[39] and the nature of diminishing returns in model improvements[40]. All analytical methods have limits to their precision and accuracy.

The natural starting point for finding solutions to new situations is to use existing models that have solved similar situations before. Deep learning frameworks like [Caffe](#) provide such facilities. For more complex applications, training new models is possible via software libraries and services like deep learning focused [Theano](#) and [DL4J](#), that simplify the creation and execution of distributed models. Advanced libraries like [TensorFlow](#) enable efficient design of deep learning model architectures. Efforts similar to [Keras](#), modularize and integrate these lower-level libraries into easily usable solutions. Machine learning libraries such as [Photon](#) [ML](#) build upon existing data platforms. Investments by companies into products like [Nervana](#) [Systems](#) [Neon](#) and [Preferred](#) [Networks](#) [Chainer](#) focus on optimizing the use of GPUs. Widely used community efforts like [scikit-learn](#) serve as the growing space for new ideas. Foundational building blocks are delivered by company products like [Nvidia](#) [Deep](#) [Learning](#) and [Microsoft](#) [DMTK](#). Advanced Analytics technologies are mostly based on Java, Scala, Python, Lua and C++. Experimental and small scale model building frequently happens in R.

When performing large scale model building, the move to machine learning platforms makes sense. There are many companies pushing the performance and capabilities of machine learning. ML platforms are developed by [Continuum](#) [Analytics](#), [H2O](#), [Dato](#), [Skytree](#) and others. There are specialized companies like manufacturing and Industrial IoT focused [Uptake](#) and others.

From a data science process perspective, generic solution companies like [DataRobot](#) are streamlining the entire process of model building, selection and deployment for organizations doing large scale model building. Some technologies and companies are specializing further in a particular step in model building like model optimization by [Sigopt](#) and deployment by [Yhat](#).

From a different angle, [Wolfram](#) is pushing mathematics-based computational knowledge, and is gaining traction in R&D intensive industries like biotechnology. Interesting IoT venues are opened by cloud HPC simulation platform [Rescale](#) that provides scalable model execution for variety of models like those by Ansys. From knowledge creation, an interesting approach is taken by [SparkBeyond](#) that is automating research as well as the search for complex patterns in data. Relevant to knowledge analytics, semantic analytics technologies are developed by companies like [Cycorp](#). Topological analysis for knowledge discovery is developed by [Ayasdi](#). The frontier of AGI is developed by companies like [Vicarious](#) and stealth startups, with a focus on applications in analyzing images and video.

As the saying goes, build what you must, buy what you can.

---

[39] G-B. Huang, Z. Bai, L. Lekamalage, C. Kasun, C. M. Vong, Local Receptive Fields Based Extreme Learning Machine, IEEE Computational Intelligence Magazine, 2015
[40] D. Ferrucci, E. Brown, J. Chu-Carroll, J. Fan, D. Gondek, A. Kalyanpur, A. Lally, J. Murdock, E. Nyberg, J. Prager, N. Schlaefer, C. Welty, Building Watson: An Overview of the DeepQA Project, AI Magazine, 2010

## Data Specific Analytics

Language, audio and vision are the most important means of communication and perceiving the world for humans. For machines to be able to interact seamlessly with people, the frontier of enabling technologies is developed by IBM Watson, Google and Microsoft, among others.

Besides bigger players, there are younger companies specializing in specific types of data analysis. For example Dextro is focused on making videos discoverable, searchable and curatable. Vision.ai is pushing vision services to the cloud. In satellite imagery, Planet Labs and others are focused on analysing the very large datasets. Clarifai Vision uses deep learning for detecting duplicates and doing visual searches. Similarly, in audio analysis there are companies like Nuance Communications providing speech and imaging applications for businesses.

Given the very large knowledge base that the written Internet represents, there are a lot of commercial activities to extract knowledge from it, see for example import.io. For building your own, high performing text parsers like SyntaxNet and industrial spaCy tokenizer provide the basis for efficient natural language processing (NLP) in English language. State of the art, research focused, NLP libraries like Carmel, NLTK, Gate provide advanced features like semantic reasoning and information extraction. For automated statistical machine translation, academic software like Moses exist. Applications of deep convolutional neural networks are the most successful in dealing with perceptual data[41].

Overall, the field of Data Specific Analytics is advancing fast and considerable progress has been made in making language, audio and vision understandable to machines.

## Usage Layer

Technologies deliver value through Applications. Here are some examples of how organizations are applying technologies in the data and IoT trend and how people can interact with the technologies.

### Applications

I explicitly list existing applications rather than communicate speculative promises. Refer to angel and venture capital investments[42] and venues like KDD to see where the market is heading:
- A frequent and most basic generic application is event and outlier detection.
- In the automotive industry, route optimization based on traffic, as demonstrated by Google Waze and Zubie, is relevant for transportation and logistics, like Uber and Didi Chuxing.
- In synthetic biology and organism engineering, Zymergen is demonstrating the use of automation, data architecture and machine learning.
- In investment finance, Alphasense and Bloomberg are leading data application frontier.
- In retail finance, Lendup, Affirm, Mirador Lending, Inventure and Earnest are applying data for better servicing their customers.
- Energy intelligence is developed by EnerNOC.
- In internal intelligence, Palantir is leading the way.
- In market intelligence look for Funderbeam, Quid, Mattermark and CB Insights.
- Bayes Impact is applying data science for non-profit initiatives.

---

[41] http://www.computervisionblog.com/2016/06/deep-learning-trends-iclr-2016.html
[42] For example Data Collective, Founders Fund, DFJ, Khosla Ventures, Andreessen Horowitz, Y Combinator.

- In personal health, companies like Jawbone, Misfit and Garmin help people track their fitness goals.
- In agriculture, the Climate Corporation has found a way to protect farmers against the climate change. John Deere is providing APIs for precision agriculture.
- In enterprise sales, Salesforce is leading the way with acquisitions like MetaMind, Implisit, MinHash and RelateIQ as well as pipeline management by InsideSales and Pipedrive. Gainsight is pushing with customer success software.
- In enterprise customer service, wise.io and NICE Systems lead to data driven decisions.
- In enterprise security and fraud detection, Feedzai and Sift Science are the frontier.
- In cloud infrastructure monitoring, Datadog.
- New generation of company wide business intelligence by companies like Looker and Domo.
- In adtech, Rocketfuel. In enterprise legal, Ravel.

There are valuable uses of data in many other areas demonstrated by many companies.

## Human Interfaces

Intelligent systems must have processes of communication and control that interface with people[43]. The report's focus is on user endpoints, rather than on data visualization and communication[44]. Machines have to adapt to the needs of people, so that they can interact naturally. User and customer experience design has become an integral part of developing applications. The most recent transformative technologies in user interfaces are virtual reality and augmented reality technologies, such as developed by HTC Vive, Magic Leap, Facebook Oculus and Meta.

In mobile technology, user interfaces like Facebook M and wit.ai, Microsoft Cortana, Google Voice Recognition, Apple Siri, Turing Robot and api.ai make interaction with devices more intuitive. Automated assistants for interaction with corporate processes, systems and people, are reducing the burden of high volume communication brought by the email, chat and mobile apps era. By 2020 it is estimated that at least 50% of all searches are going to be either through images or speech[45]. Some of the companies to follow are x.ai and ClaraBridge. Assistants in combination with solutions like IFTTT can automate and connect Internet services so that they work together for the benefit of people and make their lives feel more natural. A system moving in this direction is Viv, based on a dynamically evolving cognitive architecture extending intent of the user.

## Limitations

There are many detailed, technical limitations and API designs of existing technologies that must be taken into account in systems architecture. Here are the upper bounds for some of the factors:
- Financial cost and latency numbers[46] of storage, network and processing services.
- Geographical locations of data centers and CDNs. Physical fiber networks layout, with bandwidths realistically in the range of 10GBps on planetary scale. Coverage and quality of mobile networks. These mainly follow the trends of population and the Internet adoption.
- Linear scalability and high availability of data-center scale OS proven on 10,000s of nodes.

---

[43] T. Mikolov, A. Joulin, M. Baroni, A Roadmap towards Machine Intelligence, arXiv, 2015
[44] C. Viau, Try Datavis Now, Github Books (in preparation), 2016
[45] http://www.fastcompany.com/3035721/
[46] http://people.eecs.berkeley.edu/~rcs/research/interactive_latency.html

- Automatic scalability of Data & IoT platforms to billions of devices[47], proven scalability to trillions of messages per day[48].
- Proven scalability of storage services to a few exabytes[49]. Some organizations think in zettabytes and above. For benchmarks on data platforms see TPC[50] related materials.
- Limitations of Data Exchanges focused on a few industries like financial and geospatial.
- Resource intensity of machine learning training processes addressed with GPU and FPGA.
- Lack of good solutions for highly multidimensional data visualization and communication.

## How the Pieces Fit Together

*Illustration with a possible top down structure of a very large scale, autonomous cyber physical system concept. It is just one way of clustering and labeling a network of processes that form such a system.*

To bring different research and commercial results together I illustrate a hierarchical system of systems concept where different independent subsystems deal with different levels of abstractions. In designing the concept the approach of systems thinking is used. Systems thinking includes the idea of layered structure. Architecture is an abstract description of the entities of a system and the relationships between those entities [51]. The allocation of physical or informational elements of function into elements of form is one of the earliest and most important decisions. This vision is not intended as a technical blueprint[52], it is an illustration.

The actual systems architecture would benefit from principles like reactive systems[53], microservices[54] and patterns like the blackboard architecture to achieve non-functional requirements and provide programmable APIs to subsystems that can be used independently and for system management.

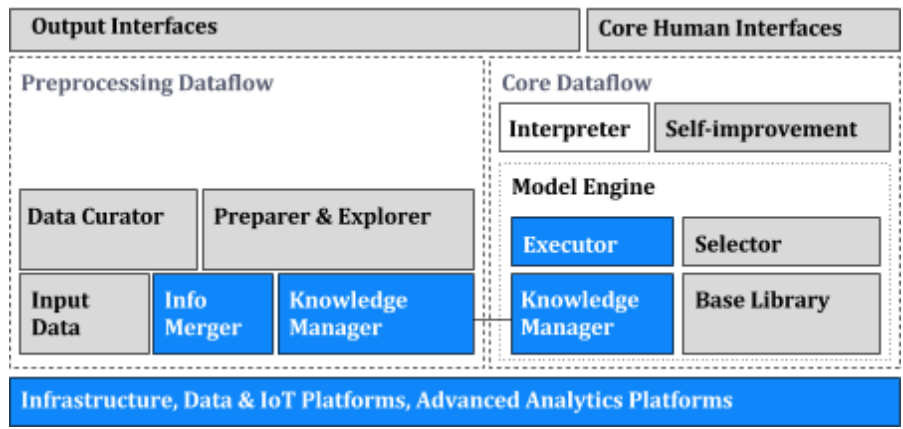

*Figure 3. A concept architecture of an intelligent system. The colors represent components with varying degree of maturity based on Technology Action: blue stands for mature, grey for emerging and white for nascent commercial grade technologies.*

---

[47] https://aws.amazon.com/iot/how-it-works/
[48] http://www.confluent.io/blog/apache-kafka-hits-1.1-trillion-messages-per-day-joins-the-4-comma-club
[49] http://www.slideshare.net/AmazonWebServices/stg306-efs-how-to-store-8-exabytes-look-good-doing-it
[50] http://www.tpc.org/
[51] E. Crawley, ESD.34 Systems Architecting — Lecture Notes. MIT Engineering Systems Division, IAP, 2007
[52] J. Spolsky, Don't Let Architecture Astronauts Scare You, Joel on Software, 2001
[53] http://www.reactivemanifesto.org/
[54] http://microservices.io/

The structure is based on a direct mapping from a more detailed Research Progress and Technology Actions analysis. Systems can be built on top of platforms such as those in the Infrastructure Layer and Processing Layer sections. There can be multiple instantiations of subsystems or the entire system.

The lowest abstraction level of the system is in Preprocessing Dataflow, that has to cope with diverse input data from other systems and its own subsystem processes. With well-prepared data, high quality features and input information, the next level of abstractions is addressed by modeling and advanced analytics focused Core Dataflow. These subsystems form the basis for intelligent systems. The system has to have a connection to the environment through well-defined Output Interfaces and Core Human Interfaces. In large systems, explicit attention to system management is necessary to help designers, developers and operators decide how to impact the overall system to evolve in a certain direction. It makes sense to observe and analyze its behaviour and interactions with explicit threat intelligence and modeling functions.

## Preprocessing Dataflow

This part of the illustration deals with lower abstraction levels of data, information and knowledge produced by both external and internal systems.

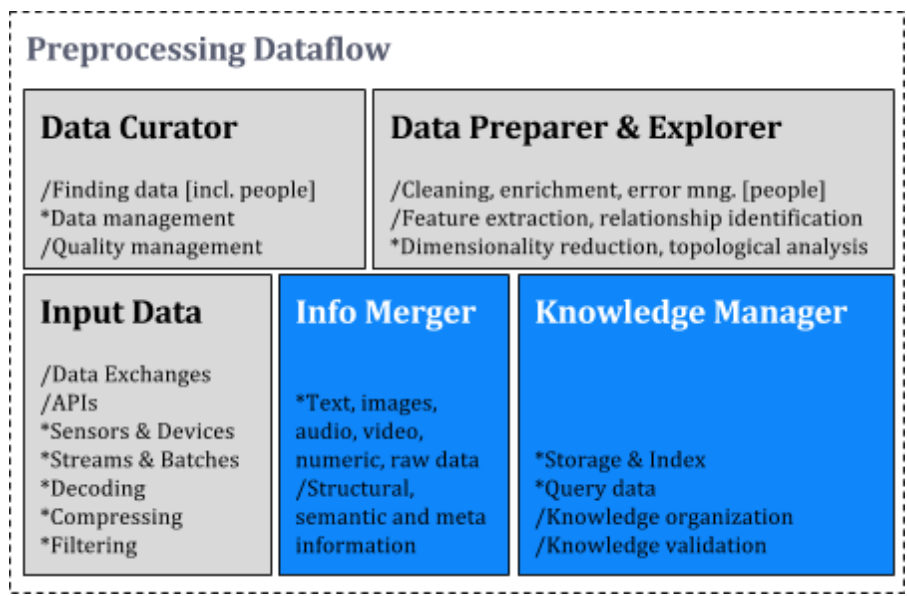

*Figure 4. Illustration of the subsystems dealing with low-abstraction levels of data, information and knowledge. * stands for a mature feature, / stands for emerging feature*

### Input Data, Info Merger and Knowledge Manager

To observe context and environments, the system has to have well-defined input APIs. Preprocessing is one of the most human resource intensive activities, so aiming for high levels of automation in this part of the system has a high return of investment.

To accelerate the progress further, the preprocessing can build upon Data Exchanges available in specific industries with high quality, preprocessed data streams and data batches. Given that the incoming information may be in a format and structure not directly usable by the system, decoding of information is necessary. As the input data is growing exponentially, the system needs measures to manage the growth rate of information. For achieving this, compression, archival and in some cases filtering and aggregation may be used.

The types of information relevant to adaptive systems vary a lot in the Input Data. To understand language, the system should be able to adapt to changes in the languages[55]. For effective processing of images, segmentation, captioning and classification is necessary. For processing video, video analytics conducting captioning and extraction of timed events from the content, is necessary. Processing audio with speech recognition can be transformed into textual representation that can be fed into next processing steps. Furthermore, both internal and external model output can provide high-dimensional input that should be classified and clustered. Vectors and multidimensional arrays are the output of this initial preprocessing step. These can be fused together based on structure, semantics and meta information, such as location, time, source, channel and destination. As shown by Data Exchanges. As a result, the Core Dataflow receives a set of unified data streams and batches for conducting advanced analytics.

For the system's ability to remember history, efficient methods of storing both structured and unstructured data, information and knowledge are necessary. The data must be indexed according to the expected queries. Efficiency in indexing is mandatory as it can be larger than the data itself. Depending on the nature of the data, there could be means for probabilistic retrieval.

## Data Curator, Data Preparer & Explorer

For ensuring the quality of input data and enabling knowledge discovery, further subsystems are required in the preprocessing part of very large scale intelligent systems.

Namely, a Data Curator has to find data through involving experts and browsing Data Exchanges. The functions for managing the data, based on its use frequency, must be implemented so that the contents in the Knowledge Manager can be grouped into hot, warm and cold data, all using different technologies with different retrieval latencies. The overall quality management function in the Data Curator should offer means for assigning a quality rank to the data as well as operate automated quality engineering processes on the data.

Input Data and the contents of the Knowledge Manager can rarely be used directly by the processes conducting advanced analytics. A Data Preparer takes care of processes that clean, normalize and do overall data enrichment before the modeling step. The functions have to deal with errors and their distributions, interpreting which requires a high degree of domain knowledge possessed by dedicated people.

As the content is high-dimensional, guidance for machines is needed to indicate what are the best features for solving a situation at hand. Feature manufacturing is a key process[56]. The Data Explorer targets automated feature extraction, feature relationship identification and dimensionality reduction processes to narrow down the sets of possible input variables for models. Given the state of technologies and the need for domain knowledge, human involvement is necessary. They can utilize new methods like topological data analysis to discover the most important relationships in the data for the models.

---

[55] http://blog.stephenwolfram.com/2015/11/how-should-we-talk-to-ais/
[56] http://www.kdnuggets.com/2015/12/harasymiv-lessons-kaggle-machine-learning.html

## Core Dataflow

This part of the illustration is responsible for taking high quality inputs from the Preprocessing Dataflow and the Knowledge Manager and turning those into metrics, estimates, predictions, simulations and projections.

The core parts of the subsystem are the Model Engine, Interpreter and Self-improvement subsystems. The Model Engine subsystems focus on providing a wide range of useful models with effective training, execution and combining. The Interpreter takes the output of the models and processes the results along with models, so that these can be effectively shared with other systems and people. The Self-improvement is about automating the processes of data science related to models.

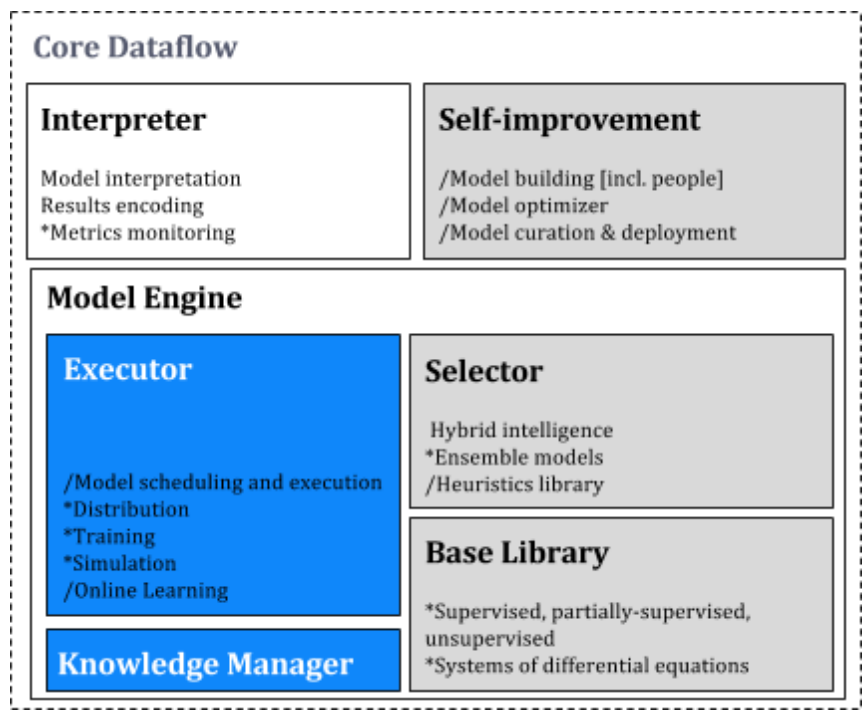

*Figure 5. Illustration of the subsystem functions in the Core Dataflow responsible for the base intelligence*

### Base Library, Executor and Selector

The Base Library consists of models as well as metadata pointers to other intelligent system APIs. The goal of the models is to find patterns, simulate specific scenarios for specific conditions, make predictions, create insights and enable decision-making. The machine learning methods and models in the Base Library can be supervised, partially-supervised and unsupervised, which means that the Self-improvement subsystem has to have means to produce and manage content in the library. Some models, especially those related to physical objects, can be models based on systems of differential equations. These require different execution mode compared to ML.

The Executor schedules and allocates resources for running selected models from a prioritized processing backlog. The distribution of the execution of training, simulation, operational and online learning models is a must-have. The Executor may break the processes into intermediate steps that may produce intermediate datasets. These can be reused for speeding up the use of models in recurring situations. Looking for patterns in the model execution steps will improve the contents of the Base Library with new models that deal with different levels of abstractions.

Successful use of models is usually a combination of methods and algorithms. The Selector builds the knowledge base of model performance on specific input combinations and uses the content for facilitating decisions when a hybrid intelligence or an ensemble model should be used. The content is managed by the Self-improvement subsystem. The Selector also holds information about the empirical resource use of models to build a library of heuristics that can serve as input for model selection.

### Self-improvement and Interpreter

Building applicable models is a process that involves people with data science and domain expertise. The processes have to be automated to a great extent to enable fast adaption to new situations and changing conditions. Model performance improvements demonstrate diminishing returns over time, thus it is necessary to be able to monitor these changes to make decisions about the maturity and applicability of solutions to different situations.

Similar to Data Curation, there should be means for model curation, as some types of methods like online learning can produce hundreds of thousands of models. Situations that are mission critical or frequently encountered by a cyber physical system, must be addressed quickly, while non-mission-critical situations can be addressed with a certain latency. This means that models can also be classified as hot, warm and cold in the libraries. This information can be used by the Executor in allocating resources and doing model deployments to runtime environment.

The output of the models, as well as the way a model produced the output, are important bits of information for sound decision-making and collaboration. For evaluating different forms of bias and uncertainties, the Interpreter collects information from the Model Engine and monitors for important occurrences of events. Since the model output, model description and metrics can be very large, this data needs to be compressed and encoded for efficient communication. This is the most speculative piece of the illustration.

## Output Interfaces and Core Human Interfaces

The events, decisions and insights are communicated through Output APIs. Defining these early in the development of a system is a must-have activity. It guides the design and provides means to define baseline subsystem implementation that can be improved over time. As with any interface that is used in real Applications, in-depth versioning and planning is required.

The Core Human Interfaces are the most important control measures next to the system management. Defining these early in the system architecture is important for success. These are also measures for safety of the entire system. For further measures of safety, there should be a chain of command built throughout the subsystems impacting system actions directly. This chain can introduce interruptibility[57] into the system.

Given the complexity of the interfaces to an Intelligent System, two main functions are proposed. One is visualization that translates data into understandable form for human decision making. The other is interactive bots with which text, audio, images and video can be exchanged. The communication with interactive bots is a high priority input channel for the Preprocessing Dataflow, introducing a direct feedback loop into the system for training models.

---

[57] http://intelligence.org/files/Interruptibility.pdf

## Notes on the Architecture

Architectural principles for systems of systems have been long explored in both industry and academia. In short, the elements must be able to usefully operate independently, have managerial independence and support evolutionary development. In communication, the interfaces have to be standardized between different layers, which allows for dedicated developments in each system. For successful software modularisation and architecture, one should assume most components will be replaced during the lifecycle of the software.

Unable to centrally analyse the data in its entirety, a system of systems remains distributed, with specific groups addressing specific aspects of the global networked system. Data protection gap needs to be explicitly addressed in the systems architecture[58]. Continuous deployment, testing and monitoring enable development speed and meeting non-functional requirements.

The traditional 7 layers of OSI model are insufficient to describe the richness of new systems. Examples are reactive systems based on the concept of Actors. Actors represent a system of completely decoupled components that interact only using (a)synchronous messages. While it can be argued that it is a part of the "applications layer", it unnecessarily bloats the layer. For future systems, such communication could be better represented by a new layer not yet present; call it the "systems layer" if you will. This layer would present systems for dynamic analysis of interaction networks to infer properties of the whole system.

## Intelligent System Initiatives

The realization of the illustrated concept would be a bottom-up endeavour by numerous dedicated parties[59]. Minimum Viable Products would be in order. Independent parties develop technologies, products and research according to their own agenda. In short, there are three types of actions relevant to such systems:

- Incremental improvement with immediate commercial impact:
    - Preprocessing Dataflow: Info Merger, Knowledge Manager
    - Core Dataflow: Executor
- Innovation and applied research with impact in 3-5 years[60]:
    - Preprocessing Dataflow: Input Data, Data Curator and Preparer & Explorer
    - Core Dataflow: Selector, Base Library and Self-improvement
    - Output Interfaces, Core Human Interfaces
- Fundamental research with tangible commercial impact in 10+ years[61]:
    - Core Dataflow: Interpreter

---

[58] http://www.gigya.com/resource/whitepaper/the-2015-state-of-consumer-privacy-personalization/
[59] S. Berkun, The Myths of Innovation, O'Reilly, 2010
[60] Based on a list of public investment sizes and rounds in companies related to the field
[61] The timeframe estimate is based on the observation that the research is emerging and ACM SigSoft Impact Project conclusion that software engineering breakthroughs take 10-15 years to reach wide scale practice.

Leading organizations who are pushing the boundaries of intelligent system developments are listed below. Smaller companies tend to keep a lower profile and are not listed. Keeping an eye on the publications, public releases and acquisitions by these companies can serve as an indicator of directions and progress in the field:

- Google
- Microsoft
- Apple
- Facebook
- AWS
- IBM
- Baidu
- IARPA
- OpenAI

It is important to release software early in order to understand the dynamics and to find improvement areas in design and implementation. Iterations and increments are essential in software engineering. Building real cyber physical systems architecture has to be based on actual requirements.

## Conclusion

The benefits of intelligent systems are demonstrated by many ANI applications used by businesses, governments and society. Research relevant to the field offers good basis for tackling complex scenarios and advancing the field. The core infrastructure like data & IoT platforms and advanced analytics platforms are maturing with potential high-value commercial applications. Relevant to the intelligent systems, the highest maturity is in data gathering and preprocessing, followed by advanced analytics.

The hierarchical system of systems concept, consisting of parts related to data preprocessing, models and self-improvement, shows how the different pieces may fit together. This vision is not intended as a technical blueprint. I estimate that the concept illustrated here is achievable within a 5-10 year timeframe.

Bringing together research and technology developments is necessary for engineering intelligent systems. Hopefully this writing sparks developments and conversations that speed up the progress.

# Appendix

Background on the Infrastructure Layer that impacts building of cyber physical systems.

## Building Blocks

The base components that drive advancements of systems are processors, memories, drives and chargers. These define the basis for non-functional requirements, especially performance.

Processors by companies like Intel[62], Qualcomm[63] and others are targeting the new IoT and data era. While the era of Moore's Law is coming to an end[64], new types of processors like Tensor Processing Units[65] and increasingly field-programmable gate arrays (FPGA) and potentially HP Nanostores[66] and adiabatic quantum optimization processors by DWave[67] [68], are pushing the boundaries of what is possible with data focused hardware. Visual computing technologies by companies like Nvidia offer step jump increases in performance of ML, as GPUs increase the bandwidth, reduce latency and communication costs with built in support for matrix algebra[69].

Mature drive technologies like Solid State Drives productized by companies like Nimble Storage, offer significant read performance improvements compared to traditional hard disk drives.

Wireless charging using new technologies by companies like uBeam and WiTricity offer new means to charge independent devices without the hassle of cables. This unlocks the potential of smart, autonomous devices in many applications.

## Devices and Machines

Devices are becoming open and extensible platforms that are capable of complex tasks independently from central control, the so-called Fog[70] and Mist Computing[71] that complement Cloud Computing. This creates new means for cyber physical systems to interact with the world.

Incumbent electronics, appliances and mobile device manufacturers like LG are continuously innovating in new types of consumer devices. At the same time new Industrial Internet machines and software are developed by giants like Siemens. Companies like NXP Semiconductors and Psikick are innovating in the foundations of circuit design suitable for IoT era. Very small electronic devices (MEMS) by companies such as mCube offer new means to interact and control devices. At the same time prototyping platforms like Arduino, Raspberry Pi and Xilinx reduce the speed and cost of innovation in creation of devices. New applications can be developed quickly.

---

[62] Intel Xeon Processor E7 Family for data analytics
[63] Qualcomm Snapdragon for mobile devices
[64] https://www.technologyreview.com/s/601102/intel-puts-the-brakes-on-moores-law/
[65] N. Jouppi, Google supercharges machine learning tasks with TPU custom chip, Google Cloud Platform Blog, 2016
[66] Ranganathan, From Microprocessors to Nanostores: Rethinking Data-Centric Systems, IEEE Computer, 2011
[67] D-Wave Quantum Hardware
[68] https://plus.google.com/+QuantumAILab/posts/DymNo8DzAYi
[69] http://www.nvidia.com/object/machine-learning.html
[70] http://www.cisco.com/c/en/us/solutions/internet-of-things/iot-fog-computing.html
[71] http://www.thinnect.com/mist-computing/

Overall commoditization and maturity of electronic components has led to major innovations of devices operating on land, air, sea and space. For example on land, products like [Google](#) [Self](#) [Driving](#) [Car](#) and [Tesla](#) are creating autonomous cars. Furthermore, incumbent companies like [Ford](#), are turning their products into open platforms. In air, companies like DJI are producing commercial and recreational unmanned aerial systems. These devices are getting information platforms for developing and operating them, for example [Airware](#) and [Dreamhammer](#). On the sea, companies like Liquid Robotics are developing autonomous ocean robots capable of covering vast distances. In space, new entrants like SpaceX and BlueOrigin are creating rockets for orbital flight. At the same time companies like [Spire](#) and [Planet](#) [Labs](#) are driving bringing down the costs of space with nanosatellites for applications like satellite imagery.

A similar transformation is happening in home automation, where home hubs like Google Nest and [Amazon Echo](#) are changing how we connect devices like thermostats and appliances at home. Companies like Tado and View Dynamic Glass are changing the historically plain devices, such as air conditioners and windows, into smart devices. Home robotics like social robots by Jibo and cleaning robots by iRobot are changing household chores and social interaction. Incumbent companies like Samsung, Honeywell and Philips are also investing heavily into home automation.

Industrial automation is evolving at a similar scale and pace. Amazon Robotics, [Starship](#) and other companies are creating mobile robotic fulfillment systems. Industrial Internet machine producers like Bosch and ABB are producing heavy-industry electronics and machinery. Younger companies like Rethink Robotics and Kuka Robotics are creating industrial robots and factory automation solutions so humans can focus on other tasks.

Increasingly more devices are attached to people with the wearables trend. Devices like Apple Watch (3 GFlops) and Pebble bring new human interfaces and ease of use to the people. Big companies like Japanese Rakuten and South Korean Samsung, and smaller ones like HumanCondition and Ringly are turning clothes, jewelry and accessories into smart endpoints.

Overall, these trends are a big enabler for builders of cyber physical systems. Innovative devices and machines enable rapid adaptation not only in software, but also on hardware side. Moravec's paradox[72] seems to be becoming less of a challenge thanks to working real-life applications.

## Operating Systems

For dealing with the diversity of devices, there is also diversity in operating systems (OS). Universal OS like Windows, Ubuntu, Debian and FreeBSD will continue to unify this layer of technologies, but the more specialized operating systems like OS X, Android, iOS, Micrium and TinyOS will have a continued role to play. This means that at least when developing applications, multiple platform support requires dedicated attention of software engineers. For universal OS, technologies like Docker simplify deployment.

When planning systems of systems, combining computations into a dynamic network of data processing quickly becomes a complex task of managing clusters of devices. Thus, new ways of expressing computation are developed by abstracting the device layer into programmable concepts, which can be combined into computational process flows. An example is the data center OS [Mesosphere](#) and [Stratoscale](#). In designing very large scale systems, it makes sense to utilize this abstraction level to increase the focus on the core functionality.

---

[72] V. Rotenberg, Moravec's Paradox: Consideration in the context of two brain hemisphere functions, Activitas Nervosa Superior, 2013

## Protocols

Different intents of communicating systems call for different protocols. This is clearly observable in the evolution of protocols in the IoT era. Most common fixed and Wireless protocols like WiFi, Bluetooth, RFID and LTE are seeing numerous additions of new industry specific protocols like Zigbee, LoRa, MQTT, NFC, 6LoWPAN, DDS, LWM2M, and application specific protocols like BitX and M-Bus. The important takeaway is that these protocols address specific needs like speed, power consumption, resource constraint, reliability, machine-to-machine communication and other aspects relevant to building cyber physical systems. Products like Eero, simplify the complexity for dedicated needs, like use at homes. The same can be expected for offices and factories.

In designing the input and output interfaces of systems we must find ways to deal with the diversity of protocols. The Data & IoT Platforms, as well as Data Exchange address this.

## Networks and Connectivity

Networks form an integral part of systems as they enable communication. The infrastructure is advancing towards open, programmable environments, capable of operating at 1GBit/s with ms latencies. Companies like Cisco, Thinnect and others continue to innovate in this arena with concepts like software defined networks and network as a service.

With the increase of machine-to machine communication, companies like Texas Instruments, Atmel and others, traditionally not focused on networks, are pushing the solution space of M2M communications with components. Younger companies like goTenna and others design and market wireless connectivity products for decentralized communications. There are also dedicated connectivity companies like Sigfox, Sierra and others pulling together different technologies to target specific needs of IoT, for example city-scale across the globe.

Large scale networking equipment and solutions are operated by telecom operators who share the strategic vision of IoT and data intensity. Thus we see companies like T-Mobile and Huawei focus on data centers. The bigger players like Verizon, AT&T, China Mobile, SoftBank, Bharti Airtel, Orange Telecom, Telefonica, Vodafone and others are all investing into their capabilities to meet the future needs of IoT and data era, where the number of devices communicating over the network is very large. The largest underdeveloped populated area of the planet is currently Africa, but with leadership by companies like Millicom and MTN Group, the network quality will increase in time. In more remote and less populated areas there is still a long way to go, but there are initiatives targeting this need as well[73].

Networked systems pose several challenges, such as variable time delays, failures, reconfigurations, not only at the lower layers of the OSI model, but at the application layer. Given that telecom operators have moved into providing applications in multimedia, we will see continued improvement in guarantees provided by the infrastructure, developed in partnership with content delivery networks (CDN) like Akamai and cloud service providers. Thus it is assumed, that in designing cyber physical systems, the designers and developers have to mainly focus on the quality of the edge communications required by the new protocols and devices.

---

[73] https://info.internet.org/

## Cloud Platforms

Cloud is the most mature concept compared to Fog and Mist Computing. Networked computation is being pushed to the edges of networks. This creates a dynamically changing infrastructure environment where functionality is continuously moving between the edges and centralized systems. Cloud services offer ways to meet many of the non-functional requirements, such as security and scalability. Initiatives like OWASP define requirements and ways for systems to advance from the base offer to more secure solutions. Companies like Ionic Security offer data protection & control platforms.

The main service providers are Google Cloud, Microsoft Azure and Amazon Web Services (AWS) who all target data and machine learning[74] [75] [76] related functionalities at planetary scale. Other providers like Salesforce Heroku can be used for fast innovation in specialized applications.

## Active Companies

A list of companies mentioned in the report. There are many more pushing the boundaries of possibilities.

| Category | Mentioned Companies |
| --- | --- |
| *Infrastructure Layer* | |
| Building Blocks | Intel, Qualcomm, Google, HP, DWave, Nvidia, Nimble Storage, uBeam, WiTricity |
| Devices and Machines | LG, Siemens, NXP Semiconductors, Psikick, mCube, Arduino, Xilinx, Google, Tesla, Ford, DJI, Airware, Dreamhammer, Liquid Robotics, SpaceX, BlueOrigin, Spire, Planet Labs, Google, Amazon, Tado, View, iRobot, Samsung, Honeywell, Philips, Bosch, ABB, Rethink Robotics, Kuka Robotics, Apple, Pebble, Rakuten, HumanCondition, Ringly |
| Operating Systems | Microsoft, Apple, Canonical, Micrium, Docker, Mesosphere, Stratoscale |
| Networks and Connectivity | Cisco, Thinnect, Texas Instruments, Atmel, goTenna, Sigfox, Sierra, T-Mobile, Huawei, Verizon, AT&T, China Mobile, SoftBank, Bharti Airtel, Orange Telecom, Telefonica, Vodafone, Millicom, MTN Group, Internet.org, Akamai |
| Cloud Platforms | Google Cloud, Microsoft Azure, Amazon Web Services, Salesforce Heroku |
| *Processing Layer* | |
| Data & IoT Platforms | Cloudera, MapR, HortonWorks, LinkedIn, Twitter, DataBricks, Data Artisans, HP, Kx, Paradigm4, Facebook, Cisco Jasper, PTC Thingworx, GE, Pivotal, Helium, Hologram, Planet OS |

---

[74] https://cloud.google.com/products/machine-learning/
[75] https://aws.amazon.com/machine-learning/
[76] https://azure.microsoft.com/en-us/services/machine-learning/

| | |
|---|---|
| Data Preprocessing and Exchanges | Tamr, Trifacta, CrowdFlower, snapLogic, Xignite, Planet OS, Enigma, Qlik, Microsoft |
| Advanced Analytics Platforms | Google, Nervana Systems, Preferred Networks, Nvidia, Microsoft, Continuum Analytics, H2O, Dato, Skytree, Uptake, DataRobot, Yhat, Sigopt, Wolfram, Rescale, SparkBeyond, Cycorp, Ayasdi, Vicarious |
| Data Specific Analytics | IBM Watson, Google, Microsoft, Dextro, Vision.ai, Planet Labs, Clarifai Vision, Nuance Communications, import.io |
| *Usage Layer* | |
| Applications | Google, Zymergen, Alphasense, Bloomberg, Lendup, Affirm, Mirador Lending, Inventure, Earnest, EnerNOC, Palantir, Funderbeam, Quid, Mattermark, CB Insights, Bayes Impact, Jawbone, Misfit, Garmin, Climate Corp, John Deere, Salesforce, InsideSales, Pipedrive, Gainsight, wise.io, NICE Systems, Feedzai, Sift Science, Datadog, Looker, Domo, Rocketfuel, Ravel |
| Human Interfaces | HTC, Magic Leap, Facebook Oculus, Meta, wit.ai, Microsoft, Apple, Turing Robot, api.ai, x.ai, ClaraBridge, IFTTT, Viv |